\documentclass[10pt, twocolumn, letterpaper]{article}

\usepackage[letterpaper,
            top=0.85in, bottom=1in,
            left=0.75in, right=0.75in,
            columnsep=0.30in]{geometry}
\usepackage{microtype}
\usepackage{amsmath, amssymb, amsthm}

\usepackage{enumitem}
\setlist[itemize]{leftmargin=1.2em, topsep=2pt, itemsep=2pt, parsep=0pt}
\setlist[enumerate]{leftmargin=1.4em, topsep=2pt, itemsep=2pt, parsep=0pt}
\usepackage{booktabs}
\usepackage{graphicx}
\usepackage{tikz}
\usetikzlibrary{arrows.meta}

\usepackage{titlesec}
\titleformat{\section}
  {\normalfont\large\bfseries}{\thesection}{0.6em}{}
\titleformat{\subsection}
  {\normalfont\normalsize\bfseries}{\thesubsection}{0.5em}{}
\titlespacing*{\section}{0pt}{1.0\baselineskip}{0.4\baselineskip}
\titlespacing*{\subsection}{0pt}{0.6\baselineskip}{0.3\baselineskip}

\usepackage[numbers,sort&compress]{natbib}
\usepackage[hidelinks]{hyperref}

\title{Bittensor Agent Arenas as a Trajectory Primitive: \\
       Distilling a Shopping Agent from ShoppingBench Subnet Traces}
\author{
  Shardul Bansal\textsuperscript{1} \quad
  Seth Schilbe\textsuperscript{1} \quad
  Jarrod Barnes\textsuperscript{2} \\[0.4em]
  \normalsize
  \textsuperscript{1}ORO AI \qquad
  \textsuperscript{2}Dynamical Systems
}
\date{\today}

\begin{document}

\twocolumn[
  \begin{@twocolumnfalse}
    \maketitle
    \begin{abstract}
      \noindent
      Small-model agentic post-training is bottlenecked less by the choice of algorithm than by the trajectory substrate it consumes. The leading recipes (RLVR, group-relative RL, rejection-sampled re-SFT) all require multi-turn traces that carry per-trajectory supervision, and the two existing sources fall short on their own: frontier-synthesised data inherits the synthesizer's biases and collapses the distribution's long tail, while unfiltered production logs are unjudged and contaminated by shortcut behaviour. We argue that an incentive-aligned agent arena can be deliberately engineered to manufacture such trajectories, and we demonstrate this on ORO Subnet 15 (SN15), a Bittensor deployment of the ShoppingBench agentic-commerce benchmark. SN15's race mechanism, LLM reasoning judge, and rotating leak-cluster-guarded problem suite together yield a corpus with three properties that separate it from both prior sources: incentive-aligned diversity, per-trajectory judging, and anti-memorised held-out evaluation. We introduce a structural-quality filter that converts the raw firehose into a trainable corpus by selecting agentic trajectories (those in which the language model itself emits the tool calls) and rejecting sub-task trajectories (those in which the language model is used only as a classifier or narrator over a deterministic search loop), and we post-train Qwen3-4B on the result with a recipe closely matched to the published ShoppingBench SFT-then-GRPO pipeline. On a leak-cluster-guarded held-out partition scored production-strict, the model lifts from the published Qwen3-4B base of $18.0\%$ ASR to $42.7\%$, a $24.7$-point gain that lands within single-problem noise of the published synthetic-data SFT-only baseline ($43.6\%$), while training on a slice that is a small fraction of a single day of subnet output. The supervised stack leaves a large pass@$8$ to pass@$1$ gap ($53.3\%$ vs $34.8\%$); a per-step teacher-grounded Dr.~GRPO reward converts that latent headroom into measurable process improvement, and we identify the sub-task firehose currently outside our training pipeline as the primary lever for closing the remaining gap to the published $48.7\%$ SFT+GRPO bar. We release the filter, the corpus splits, and the arena mechanics.
    \end{abstract}
    \vspace{0.5\baselineskip}
  \end{@twocolumnfalse}
]


\section{Introduction}
\label{sec:introduction}

%

\begin{figure*}[t]
\centering
\begin{tikzpicture}[
    box/.style={
        rectangle, draw=black!70, fill=white,
        minimum width=3.7cm, minimum height=3.0cm,
        align=center, rounded corners=3pt,
        font=\small
    },
    arrow/.style={->, >={Stealth[length=3.5mm]}, thick, draw=black!65},
    feedback/.style={->, >={Stealth[length=3.5mm]}, thick, dashed, draw=black!50}
]

\node[box] (arena) at (0, 0) {
    \textbf{SN15 Agent Arena}\\[0.4em]
    \footnotesize Independent miners\\
    \footnotesize submit AI agents\\
    \footnotesize to a continuous race\\
    \footnotesize on ShoppingBench.\\
    \footnotesize Every trajectory is\\
    \footnotesize scored by an LLM\\
    \footnotesize reasoning judge.
};

\node[box] (pipeline) at (4.5, 0) {
    \textbf{Data Pipeline}\\[0.4em]
    \footnotesize Trac format\\
    \footnotesize normalisation,\\
    \footnotesize judge-coefficient gate,\\
    \footnotesize structural-quality\\
    \footnotesize filter, leak-cluster\\
    \footnotesize held-out split.
};

\node[box] (train) at (9.0, 0) {
    \textbf{Policy Training}\\[0.4em]
    \footnotesize Qwen3-4B base.\\
    \footnotesize SFT $\to$ rejection-\\
    \footnotesize sampled re-SFT $\to$\\
    \footnotesize continued teacher-SFT\\
    \footnotesize $\to$ KTO $\to$ Dr.\,GRPO\\
    \footnotesize turn-level RL.
};

\node[box] (model) at (13.5, 0) {
    \textbf{Distilled Agent}\\[0.4em]
    \footnotesize Held-out ASR lifts\\
    \footnotesize $18.0\%\!\to\!\mathbf{42.7\%}$ on\\
    \footnotesize the leak-cluster-guarded\\
    \footnotesize 75-problem partition.\\[0.3em]
    \footnotesize Re-enters the arena\\
    \footnotesize as a miner.
};

\draw[arrow] (arena.east)    -- (pipeline.west);
\draw[arrow] (pipeline.east) -- (train.west);
\draw[arrow] (train.east)    -- (model.west);

\draw[feedback] (model.south) -- ++(0, -0.6) -| (arena.south);

\end{tikzpicture}
\caption{End-to-end pipeline of this work. Trajectories produced by an incentive-aligned agent arena (left) flow through a data pipeline that normalises, judges, and structurally filters them (centre-left) to yield a trainable corpus, on which Qwen3-4B is post-trained via a five-stage recipe (centre-right) into a distilled shopping agent (right). The dashed arrow indicates the substrate-level feedback loop: the distilled agent can re-enter the arena as a competing miner, raising the bar for the next generation of trajectories.}
\label{fig:pipeline}
\end{figure*}
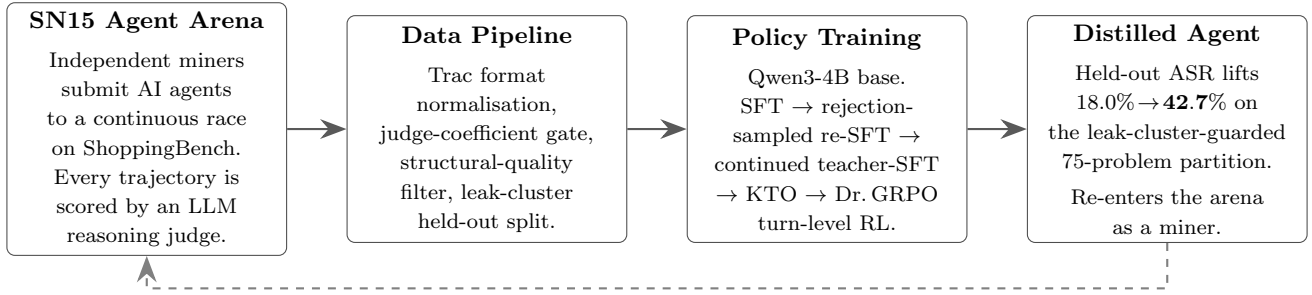

Recent work on tool-using and web-browsing agents has converged on a recurring observation: as base model capability climbs, the marginal lift on agentic benchmarks (SWE-Bench, OSWorld, WebArena, ShoppingBench) comes increasingly from the post-training recipe applied to the agent, that is, the choice of algorithm, the verifier signal, and the trajectory filter at each stage, rather than from raw parameter count or pre-training scale \citep{jimenez2024swebench, zhou2024webarena, xie2024osworld, shoppingbench2025}. The recipes that lead the recent literature, including RLVR, group-relative RL with external verifiers, and rejection-sampled re-SFT, share a structural feature: they consume trajectories that carry per-trajectory supervision independent of the synthesizer, and they pay back the additional supervision in measurable lift. A capable shopping agent has to issue many short tool calls, reason over noisy product listings, track soft constraints across turns, and recover from dead ends, all before it commits to a recommendation. The supervised and reinforcement signals required to train such behaviour are far harder to source than ordinary instruction data: a single problem produces a tree of partially correct trajectories, only a small fraction of which represent a policy worth imitating, and the difference between the good and the bad is rarely surface visible. The post-training recipe is therefore only as strong as the trajectory substrate it has access to, and the substrate question is increasingly about per-trajectory judged quality rather than sheer volume.

Two families of data have been used to fill this gap, and each has well-documented limitations when used on its own. The first is fully synthetic data, where a frontier model is prompted to roleplay both the user and the agent and the resulting traces are filtered with hand-written rules. This is the recipe used by the original ShoppingBench paper \citep{shoppingbench2025} and by widely-cited tool-use corpora such as ToolLLM and ToolACE \citep{qin2023toolllm, liu2024toolace}, and it has produced strong open baselines. The recipe inherits the synthesizer's biases, however, and the broader literature on training language models on recursively generated outputs has shown that pure self-distillation reliably collapses the long tail of the data distribution before it degrades aggregate quality \citep{shumailov2024collapse, xu2024magpie}. The recipes that have escaped this trap, such as T{\"u}lu 3's RLVR and DeepSeek-R1's verifier-grounded RL \citep{lambert2024tulu3, deepseek2025r1}, all share a property the original ShoppingBench recipe lacks: a verifier external to the synthesizer that grades each trajectory before it enters the training set. The second family is unfiltered production logs from deployed agents. These contain real diversity but are unjudged, unbalanced, dominated by the prevailing top deployed policy, and routinely contaminated by shortcut behaviour in which an agent produces the right answer through illegitimate reasoning. Neither family, on its own, is the clean source of judged multi-turn trajectories that small-model agentic post-training appears to require.

\paragraph{Building a subnet to manufacture trajectories.}
The contribution of this work is to argue, and to demonstrate, that an incentive-aligned agent arena can be deliberately engineered as a trajectory-data generator, and that the resulting corpus is the right shape for small-model agentic post-training. We built ORO Subnet 15 (SN15) on Bittensor with exactly this goal. We designed its task surface, scoring pipeline, and rotation schedule so that the trajectories falling out of it have three properties that, taken together, separate them from both synthetic data and ordinary production logs, and that we describe in detail in Section~\ref{sec:arena}: \textbf{(i) incentive-aligned diversity}, with many independent teams continuously paid in tokens to discover policies their competitors have not; \textbf{(ii) per-trajectory judging}, by an LLM reasoning judge we built that scores each trajectory's reasoning quality alongside its outcome; and \textbf{(iii) rotating held-out problems}, with a leak-cluster guard that defeats paraphrase memorisation. The novelty is not that any single property is unprecedented; it is that the whole combination is produced as a continuous side effect of a single economic mechanism we built and operate. The diversity property in particular is a property of the \emph{firehose}, not of every slice of it: the trained model in this paper draws from the agentic subset of the firehose, which inherits properties (ii) and (iii) directly but is a narrower draw from (i) than the firehose as a whole, and Section~\ref{sec:future} returns to bringing the rest of the firehose into training as the substrate-side lever.

\paragraph{ShoppingBench as a concrete instance.}
We instantiate this argument on the ORO Subnet 15 (SN15) deployment of ShoppingBench, a public benchmark in which agents are given a natural-language shopping intent and are scored on whether their final recommendation satisfies the intent's price, service, SKU, and attribute constraints. SN15 has produced a large standing corpus of judged trajectories on this task surface, governed by a race system, a reasoning judge, and a versioned problem suite rotation that together implement the three properties above. We treat this corpus, not the underlying agents, as the contribution of the subnet to the broader research community.

\paragraph{Distilling a small open agent.}
On the model side, we ask a focused question: can the structurally-filtered slice of this corpus post-train a small open model into a competent ShoppingBench agent using a recipe closely matched to the published ShoppingBench SFT-then-GRPO pipeline? We start from the same base model family (Qwen3-4B), apply the same broad two-stage pipeline (supervised fine-tuning on filtered trajectories, followed by reinforcement learning with tool-grounded rewards on multi-turn rollouts), and evaluate on a leak-cluster-guarded held-out partition of the same task surface. We make two documented substitutions against the published recipe (continued teacher-SFT in place of step-wise on-policy distillation, and KTO preference refinement as an added stage); both are motivated in Section~\ref{sec:training}. The first-order difference between our setup and the published baseline is then the source of the trajectories: subnet-generated and judge-filtered, instead of synthesised by a frontier model.

\paragraph{Contributions.}
This paper makes four contributions:

\begin{itemize}
    \item We characterise Bittensor agent subnets as a new shape of training-data primitive for agentic post-training, identifying incentive-aligned diversity, per-trajectory judging, and rotating held-out problems as the structural properties that make subnet traces qualitatively different from both synthetic data and ordinary production logs.
    \item We document the SN15 arena mechanics in enough detail to make these properties reproducible, including the race and qualifying loop, the LLM reasoning judge, and the versioned problem suite rotation, and we surface the trajectory firehose they collectively produce.
    \item We propose a structural-quality filter for converting raw subnet trajectories into a trainable SFT corpus. The filter applies a judge-coefficient gate, a format-validity gate, a deduplication and structural-quality ranking pass, and a final selection step that keeps agentic trajectories and rejects sub-task trajectories. We release the filter as code along with the trainable corpus it produces.
    \item We post-train Qwen3-4B on this corpus with a recipe closely matched to the published ShoppingBench SFT-then-GRPO pipeline (with two documented substitutions detailed in Section~\ref{sec:training}) and report results on a leak-cluster-guarded held-out partition, framed against both the published Qwen3-4B base ($18.0\%$) and the published SFT+GRPO bar ($48.7\%$). Full numbers and per-component breakdowns are reported in Section~\ref{sec:results}.
\end{itemize}

\paragraph{What this paper does not claim.}
We do not claim that subnet trajectories can replace either synthetic data or production logs in general; for many task surfaces neither subnets nor the surrounding evaluation infrastructure exist yet. We do not claim that the resulting model is a finished product. And we are explicit, in Section~\ref{sec:limitations}, about three risks that the reader should keep in mind throughout: the skew toward sub-task trajectories in the raw firehose (much of which is filtered out before training), the sensitivity of measured lifts to harness format and prompting nudges, and the fact that the supervision available from a reasoning judge is correlational rather than gold.

%

\section{Related work}
\label{sec:related}

We compare against three research threads: post-training corpora for tool-using agents, training and evaluation of commerce-specific agents, and incentive-aligned data generation. The comparison axes are the three structural properties claimed in Section~\ref{sec:introduction}: source of diversity, per-trajectory verifier, and treatment of held-out evaluation.

\paragraph{Agentic post-training corpora.}
ToolLLM and ToolBench \citep{qin2023toolllm} introduced large-scale API tool-use SFT data generated by frontier-model role-play; ToolACE \citep{liu2024toolace} adds a self-evolution loop, and Magpie \citep{xu2024magpie} generalises the recipe to alignment data. T\"ulu 3 \citep{lambert2024tulu3} and DeepSeek-R1 \citep{deepseek2025r1} demonstrate that recipes adding an external verifier to synthetic data materially outperform ones that do not. All five generate the corpus as a one-shot batch, attach supervision to outcomes rather than per-trajectory reasoning, and construct held-out partitions offline. The model-collapse literature \citep{shumailov2024collapse} is the published reason why pure self-distillation is brittle at scale.

\paragraph{Production-trajectory post-training in vertical coding agents.}
The closest production-side precedent is the new generation of vertical coding-agent companies. Cursor's Composer 2.5 \citep{cursor2025composer} post-trains a strong open base with targeted SFT and on-policy RL against rollouts in a controlled coding environment, feeding successful RL trajectories back into the SFT pool. Cognition's SWE-1.5 \citep{cognition2025swe15} reports a closely matched recipe with three grading mechanisms (test suites, code-quality rubrics, and agentic grading via browser-based testing) in place of a single outcome score. Both are explicit that production trajectories, not synthetic data, are what makes the recipe land. Our work is the agentic-commerce instantiation of the same playbook: trajectories from a continuous subnet competition rather than a single company's product telemetry, graded by a per-trajectory LLM reasoning judge rather than a stack of outcome rubrics.

\paragraph{Commerce and shopping agents.}
ShoppingBench \citep{shoppingbench2025} established the Qwen3-4B SFT-then-GRPO recipe and a synthetic GPT-4.1 trajectory corpus that we measure ourselves against. The Shopify Flow engineering write-up reports a matched recipe (synthetic trajectories plus an LLM judge) in production and explicitly acknowledges that offline benchmarks mask a substantial fraction of real-world failure modes. Amazon Rufus has been described in public communications as a retrieval-augmented shopping assistant, but no methodology is published. The consistent gap across the published commerce work is the absence of a continuously refreshed, judged, held-out evaluation surface, which is precisely the surface SN15 produces as a side effect of its race mechanism.

\paragraph{Incentivised and decentralised data generation.}
LMSYS Chatbot Arena \citep{zheng2024chatbot} crowdsources continuous head-to-head preference judgements, but the supervision is a pairwise preference rather than a multi-turn judged trajectory, and the population is uncompensated rather than incentive-aligned, which compresses diversity as the most active users self-select. Within the Bittensor ecosystem \citep{bittensor2024whitepaper}, sibling subnets run continuous competitions on different task surfaces (inference-cost minimisation, pretraining races, crowdsourced annotation) and produce data shapes that are not directly usable as agent trajectories. To our knowledge, the specific combination we describe in Section~\ref{sec:arena} (a multi-turn agentic task surface, a per-trajectory LLM reasoning judge, rotating held-out problems, and continuous token emissions tied to the consensus ranking) has not been described in the public literature.

%

\section{Background}
\label{sec:background}

\paragraph{Bittensor primitives.}
Bittensor \citep{bittensor2024whitepaper} is a permissionless network of independent \textit{subnets}, each running its own continuous evaluation competition. Three roles matter for this paper. A \textbf{miner} is an independent participant who submits a candidate model or agent to the subnet for evaluation. A \textbf{validator} is an independent operator who runs miner submissions against the subnet's task surface, scores the outputs, and reports the scores back to the chain. \textbf{Emissions} are the token reward stream the chain distributes to miners at each block, with the per-miner share determined by the validator consensus over recent scores. Every interaction with the subnet, both submission and evaluation, is cryptographically signed by an SS58 \textbf{hotkey}, which makes every artefact stored downstream traceable to the participant that produced it.

\paragraph{ShoppingBench task surface.}
ShoppingBench \citep{shoppingbench2025} is a multi-turn tool-use benchmark for agentic commerce. The agent receives a natural-language shopping intent (for example, ``find a black hosport waist bag under twenty dollars from a shop with at least 4.5 stars'') and is required to issue tool calls against an indexed product catalog until it commits to a final recommendation. The tool surface used in this paper, which is a small superset of the surface in the published benchmark, exposes catalogue-lookup tools (search by query, fetch product by identifier, fetch shop information, fetch voucher information), three helper tools we added to support multi-constraint queries (attribute-match check, in-shop product expansion, voucher arithmetic), and two terminal tools (\texttt{recommend\_product} and \texttt{terminate}).\footnote{Paper-upstream ShoppingBench at \texttt{yjwjy/ShoppingBench} exposes the catalogue-lookup and terminal tools only; the three helpers are added in our fork at \texttt{ORO-AI/ShoppingBench}. The fork is what SN15 evaluates against and what we use for all numbers in this paper.} A valid trajectory ends with exactly one \texttt{recommend\_product} call followed by \texttt{terminate}; trajectories that terminate without a recommendation are scored as failures.

\paragraph{ASR and the four rule axes.}
The headline metric reported in the ShoppingBench paper is the \textbf{average success rate} (ASR), defined as the fraction of held-out problems on which the agent's final recommendation simultaneously satisfies all of the problem's rule constraints. Rule constraints are decomposed into four axes: \textbf{price} (the recommended product's price falls within the requested band), \textbf{service} (the recommended product ships under the requested service terms, for example free shipping or cash on delivery), \textbf{SKU} (the recommended product carries the requested SKU-level attributes), and \textbf{attribute} (the recommended product matches the requested free-form product attributes, for example colour, material, or feature). We report ASR as the headline number, alongside the per-axis breakdown, throughout the paper.

\paragraph{Index-based agentic commerce.}
ShoppingBench is an example of \textbf{index-based} agentic commerce, in which the agent operates against a static, queryable product index rather than against a live retail website. The substrate is shared by industry shopping agents that route through structured catalogues (notably the published Shopify Flow recipe) and contrasts with browser-use agents that navigate live web pages (Computer Use, Operator). The choice of substrate is load-bearing for both the task and the evaluation. Index-based evaluation is reproducible by construction (the same catalogue snapshot returns the same observations for any agent), it is deterministic (no transient network state can leak into the score), and it scales to the trajectory volumes a competitive subnet requires. SN15 inherits this substrate from ShoppingBench, and the rest of the paper assumes that all evaluation runs through the same indexed catalogue.

%

\section{SN15 arena mechanics}
\label{sec:arena}

SN15 is engineered, end to end, so that the three trajectory properties claimed in Section~\ref{sec:introduction} arise from the mechanism rather than from luck. The Bittensor primitives the subnet runs on (miner, validator, emissions, hotkey-signed submissions) are introduced in Section~\ref{sec:background}; SN15 additionally gates every submission with a static-analysis pass that rejects hard-coded answer tables and a small set of restricted constructs, and persists every validator evaluation append-only. The rest of this section names the three mechanisms that make this baseline produce trainable data.

\paragraph{Incentive-aligned diversity via the race.}
Submitted agents are filtered through a qualifying phase, then enter a continuous race against a rotating problem bank. The race promotes a new top agent on a daily cadence, rate-limits each miner to one submission every twelve hours, and embargoes the winning miner's source code until the next weekday noon Pacific. The daily promotion, the cooldown, and the embargo together sustain a discover-then-share pressure that no single synthesizer can replicate: many independent teams are continuously paid to find policies their competitors have not yet found.

\paragraph{Per-trajectory judging.}
Every race trajectory is scored along two independent axes. An outcome score is computed by a fixed rule-based scorer on the four ShoppingBench rule axes (price, service, SKU, attribute). A reasoning-quality coefficient is then assigned by an LLM judge we built, which reads the full think-and-tool-call trace and assigns a coefficient that multiplies the outcome score before aggregation. The judge is trace-grounded (it must cite specific spans of the trace as evidence), decoupled from the outcome scorer, and run by validators rather than miners. The reasoning coefficient is the per-trajectory process-supervision signal that synthetic pipelines have to manufacture and that production logs do not have at all.

\paragraph{Rotating held-out problems.}
The race operates against two pools. Three small versioned suites (v1, v2, v3) are reserved as untouched evaluation territory and never enter training. A continuously growing race bank is rotated through open and held-out partitions on a cadence we control, with a leak-cluster guard that groups paraphrased or attribute-reordered variants of the same underlying problem so that surface-form drift cannot smuggle held-out problems into the open pool. The combination is what makes held-out performance a real generalisation surface rather than a paraphrase trap.

\paragraph{Trajectory provenance.}
Each evaluation writes a single canonical trajectory record carrying the agent version and miner hotkey, the validator hotkey, the problem identifier, the full multi-turn trace, the outcome score and its rule-axis breakdown, the reasoning coefficient and the trace spans the judge cited as evidence, and the scoring-pipeline version. Every row that enters the training corpus in Section~\ref{sec:corpus} can therefore be traced back to a specific agent, validator, and chain commit, and every row carries the reasoning coefficient as a separate column from the outcome score, which is what makes the structural filter in Section~\ref{sec:corpus-filter} possible.

%

\section{Trajectory corpus and structural filter}
\label{sec:corpus}

We convert the SN15 trajectory firehose described in Section~\ref{sec:arena} into a training corpus in two steps. First, every raw trajectory is normalised into a canonical OpenAI tool-calls schema using Trac, an annotation framework contributed by Jarrod Barnes (Dynamical Systems);\footnote{Open-sourced at \texttt{jbarnes850/Trac}.} we use only Trac's converter front-end in production (see Appendix~\ref{app:trac-stages}). Second, we apply a structural-quality filter to the normalised traces.

\paragraph{Structural-quality filter.}
\label{sec:corpus-filter}
The filter has three stages, detailed in Appendices~\ref{app:format-gate} and \ref{app:ranking-signals}. A \textbf{reasoning-coefficient gate} drops any trajectory whose validator-side reasoning-quality coefficient (Section~\ref{sec:arena}) falls below a fixed threshold; this is the concrete point at which the per-trajectory judge becomes load-bearing for training, in that it determines which trajectories are even eligible to enter the structural filter. A \textbf{format-validity gate} then hard-rejects malformed traces on five structural invariants. A \textbf{deduplication and ranking pass} picks one to two trajectories per problem by a small set of structural-quality signals (think-and-tool-call depth, search-query reformulation count, presence of a verification step, step-shape regularity). The reasoning coefficient gates the pool; the structural signals rank within it. A final \textbf{axis-A versus axis-B split} retains only the agentic trajectories: axis-A trajectories are those in which the language model itself emits the tool calls and the surrounding Python harness only dispatches them, while axis-B trajectories are those in which the surrounding Python emits tool calls and the language model is used only as a classifier, scorer, or narrator. A model fine-tuned on axis-B traces becomes a strong intent classifier, match scorer, and narrator, but does not become a competent agent because there is no agent-level policy in the data to imitate. Section~\ref{sec:limitations} returns to the axis-A versus axis-B distinction as the dominant constraint on the headline number, and Section~\ref{sec:future} sketches how the axis-B firehose is brought into training in follow-up work.

\paragraph{Leak-cluster-guarded held-out partition.}
\label{sec:corpus-heldout}
A naive train-eval split at the trajectory level would silently leak. Race-bank problems are answered repeatedly across rotations and miner generations, with surface-text differences (paraphrase, attribute reordering, retailer substitution) that look like distinct problems but are not, and the leak is invisible to outcome metrics. We construct the held-out partition with a leak-cluster guard (Appendix~\ref{app:heldout-construction}): leak clusters are partitioned into a training pool, an evaluation pool, and a never-touch pool that reserves the static problem suites used by the subnet's formal miner evaluation; the intersection between training and evaluation problem identifiers is verified to be exactly empty across leak clusters; and the evaluation pool is stratified by intent bucket (product, shop, voucher), seeded for reproducibility, and frozen before any training kicks off. With the guard in place, the held-out partition reported in Section~\ref{sec:results} is a real generalisation surface within the task family.

%

\section{Post-training pipeline}
\label{sec:training}

The pipeline proceeds in five stages on top of the Qwen3-4B base and the structural-filter corpus from Section~\ref{sec:corpus}, with two substitutions against the published recipe \citep{shoppingbench2025} that we motivate below. Hyperparameters, the practice-run bug catalog, and the renderer-parity work live in Appendix~\ref{app:training}.

\paragraph{Stage 1: SFT on the structural-filter corpus.}
\label{sec:training-sft}
One-epoch LoRA SFT on the structural-filter corpus produces the SFT base. The reported runs use the shared SFT configuration in Appendix~\ref{app:sft-config}.

\paragraph{Stage 2: Rejection-sampled re-SFT.}
\label{sec:training-rejection}
We sample $k = 8$ rollouts per training-set problem at temperature $0.9$, score each under the ShoppingBench bucket-aware reward, keep only score-$1.0$ rollouts, and concatenate with the SFT corpus before re-fine-tuning. The leak-cluster guard from Section~\ref{sec:corpus-heldout} is enforced at sampling time so that no held-out problem enters the training pool. Detailed parameters in Appendix~\ref{app:rejection-config}.

\paragraph{Stage 3: Continued SFT on frontier-teacher rollouts (substitutes step-wise OPD).}
\label{sec:training-teacher}
The published recipe runs step-wise on-policy distillation against a frontier teacher at this point. Per-step KL targeting over multi-turn rollouts at the corpus scales we operate at is prohibitively expensive, so we substitute continued SFT on a Sonnet 4.6 teacher's production-strict-scored rollouts through the same evaluation harness. The substitution trades the per-token KL signal of step-wise OPD for an outcome-filtered corpus, which is sufficient for the post-training lift we report. Corpus details in Appendix~\ref{app:teacher-config}.

\paragraph{Stage 4: KTO preference refinement.}
\label{sec:training-kto}
We use KTO \citep{ethayarajh2024kto} for the preference-refinement pass. KTO accepts unpaired desirable and undesirable labels, which fits a setting in which the undesirable trajectories share the format of the desirable ones but differ on rule-check satisfaction. The desirable and undesirable labels come from production-strict scoring of the teacher trajectories used in Stage 3. On the parity-validated held-out evaluation (Figure~\ref{fig:perbucket75}), KTO does not produce overall lift over the post-teacher-SFT checkpoint: both score $32$ of $75$ problems. Within buckets, KTO redistributes the policy (voucher $+4.3$pp, product $-3.6$pp, shop flat); we read this as the refinement moving probability mass within buckets without expanding the underlying capability set. A tighter-$\beta$ variant over-optimised and regressed across every bucket, which we report as a negative result. We retain the post-teacher-SFT checkpoint as the final reported model. Configuration in Appendix~\ref{app:kto-config}.

\paragraph{Stage 5: Dr.~GRPO turn-level RL.}
\label{sec:training-grpo}
The final stage applies a Dr.~GRPO \citep{shao2024deepseekmath, liu2025drgrpo} turn-level reinforcement-learning pass on top of the post-teacher-SFT checkpoint. The structural motivation is that rejection sampling (Stage 2) and KTO (Stage 4) only extract signal from groups in which at least one rollout succeeds, with the gradient applied uniformly across the tokens of a kept trajectory. Dr.~GRPO extracts signal from every rollout in the group through a group-relative advantage, applies the gradient per-token, and removes the group-standard-deviation normalisation that suppresses the signal on saturated groups. It is the only stage that can apply a negative gradient to systematic failure modes and per-token credit to the residual pass@$k$ gap.

We attempted three Dr.~GRPO variants on top of the post-teacher-SFT checkpoint, and the progression between them is informative for downstream work. The first variant (outcome-only reward, small $B$, low KL anchor) plateaued at $40$-$43\%$ over the first twenty optimisation steps; diagnostic logging traced this to optimisation instability rather than to reward degeneracy. The second variant (stability fix on $B$, learning rate, and $\beta$; same reward) contained the KL spikes but the mean reward did not climb through the first ten steps. We read this as the sparse outcome-only reward failing to provide a usable gradient at the step granularity Dr.~GRPO operates at.

The third variant followed the ShoppingBench paper's own recipe and replaced the outcome-only reward with a blended outcome-plus-tool-match reward, in which the tool-match component grades the policy's per-step tool-name and tool-argument choice against a Sonnet 4.6 teacher trajectory. We retained the stability hyperparameters from the second variant and raised the per-rollout step cap to match the held-out evaluation. The third variant trained stably for twenty optimisation steps under a clean wall-clock limit and produced a clear \emph{process} climb: the in-training rule score lifted from $0.02$ at step one to $0.42$ at step twenty, the per-step product-ID hallucination count fell from fourteen to zero, and partial successes rose from zero to twenty-four of forty-eight on a held-out diagnostic. The strict exact-match success rate, however, remained at zero across all twenty steps; the policy learned to navigate toward the correct catalogue neighbourhood without yet learning to land on the exact ground-truth product. The curve was still climbing. Full hyperparameters for all three variants in Appendix~\ref{app:grpo-variants}.

The post-teacher-SFT checkpoint has a measurable pass@$k$ headroom on the held-out partition, with pass@$8$ at $53.3\%$ versus pass@$1$ at $34.8\%$, which says that the latent capability for a downstream stage to extract is present. The third Dr.~GRPO variant converts that latent capability into process improvement, and the trend across the three variants identifies the right shape of reward signal: dense, per-step, teacher-grounded. Closing the remaining gap to exact-match conversion is the open frontier (Section~\ref{sec:future}).

%

\section{Results}
\label{sec:results}

\begin{figure}[tb]
    \centering
    \includegraphics[width=\columnwidth]{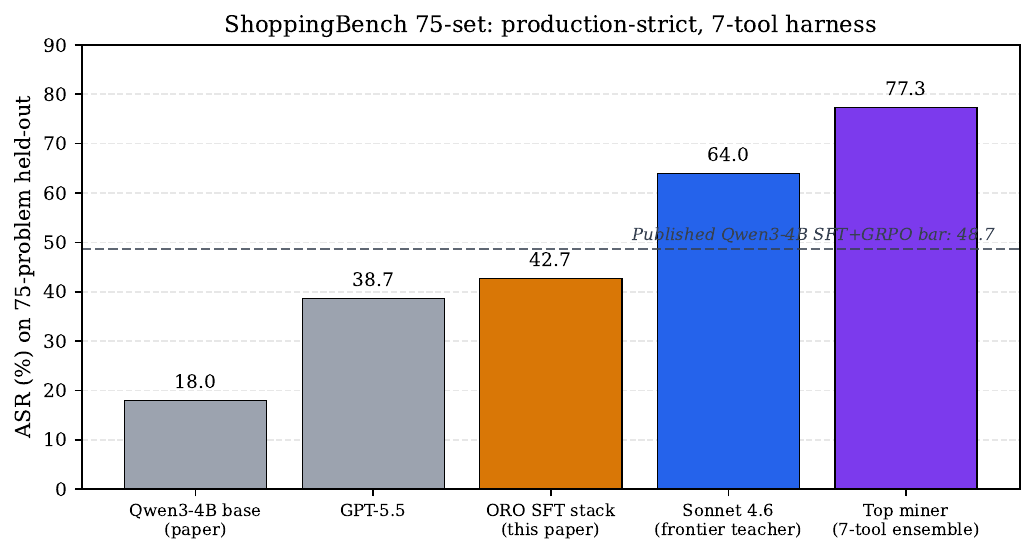}
    \caption{Overall ASR on the 75-problem leak-cluster-guarded held-out partition, scored production-strict through the 7-tool ORO ShoppingBench harness. The ORO SFT stack at 42.7\% sits roughly halfway between GPT-5.5 (38.7\%) and the frontier teacher (Sonnet 4.6, 64.0\%), and within single-problem noise of the published Qwen3-4B SFT-only baseline (43.6\% from the ShoppingBench paper) despite using subnet trajectories rather than synthetic ones as the SFT corpus. The dashed reference line at 48.7\% is the published Qwen3-4B SFT+GRPO bar; closing the gap to that line is identified as the open frontier in Section~\ref{sec:training-grpo}.}
    \label{fig:overall75}
\end{figure}

\begin{figure}[tb]
    \centering
    \includegraphics[width=\columnwidth]{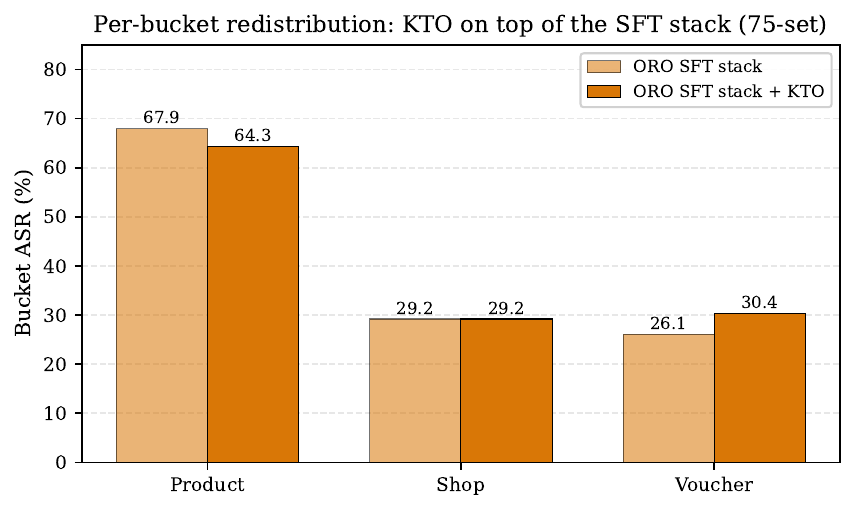}
    \caption{Per-bucket effect of the KTO refinement (Section~\ref{sec:training-kto}) on top of the SFT stack, on the 75-problem held-out. Overall ASR is unchanged at 32 of 75 problems for both checkpoints. KTO redistributes probability mass within buckets: voucher ASR rises by 4.3 percentage points, product ASR falls by 3.6, and shop ASR is flat. We read this as bucket-level redistribution without expansion of the underlying capability set.}
    \label{fig:perbucket75}
\end{figure}

We report the SFT stack described in Section~\ref{sec:training} as the final model. Figure~\ref{fig:overall75} shows the headline ASR comparison on the 75-problem held-out, Figure~\ref{fig:perbucket75} shows the per-bucket effect of the KTO refinement, and Table~\ref{tab:variants} consolidates every variant worth discussing into a single comparison.

\begin{table*}[t]
    \centering
    \small
    \caption{All variants worth discussing in this paper. Production-strict scoring through the 7-tool ORO ShoppingBench harness on the 75-problem leak-cluster-guarded held-out, unless otherwise noted in the Notes column.}
    \label{tab:variants}
    \begin{tabular}{lcl}
    \toprule
    \textbf{Model / variant} & \textbf{75-set ASR} & \textbf{Notes} \\
    \midrule
    \multicolumn{3}{l}{\emph{Published baselines from the ShoppingBench paper}} \\
    Qwen3-4B base                              & 18.0\%  & Paper baseline \\
    Qwen3-4B + paper SFT (synthetic)           & 43.6\%  & Paper recipe with synthetic GPT-4.1 trajectories \\
    Qwen3-4B + paper SFT+GRPO                  & 48.7\%  & Paper bar this work is measured against \\
    \midrule
    \multicolumn{3}{l}{\emph{Frontier closed-source baselines}} \\
    GPT-5.5                                    & 38.7\%  & 90-set: 34.4\% \\
    Claude Sonnet 4.6                          & 64.0\%  & Frontier teacher; sourced rollouts for Stage 3 \\
    \midrule
    \multicolumn{3}{l}{\emph{ORO SFT stack progression (this paper)}} \\
    \texttt{rejection\_v2}                                & 34.7\%  & Rejection-sampled re-SFT; 90-set: 25.6\% \\
    \texttt{opd\_sonnet}                                  & 34.7\%  & Continued teacher-SFT; 90-set: 33.3\% \\
    \texttt{opd\_rebal}                                   & 40.0\%  & Voucher-balanced teacher corpus; 90-set: 33.3\% \\
    \textbf{\texttt{opd\_renderers}} (final SFT stack)    & \textbf{42.7\%} & Final reported model; 90-set: 35.6\%; pass@1 34.8\% / pass@8 53.3\% \\
    \hphantom{XX}+ KTO ($\beta=0.02$, 1 ep)               & 42.7\%  & No net lift on parity eval; bucket-level redistribution \\
    \hphantom{XX}+ KTO ($\beta=0.01$, 2 ep)               & 35.1\%  & Negative result; over-optimised \\
    \hphantom{XX}+ Dr.~GRPO v19 (per-step tool-match)     & process climb & rule\_score $0.02 \to 0.42$; hallucinations $14 \to 0$; exact-match unchanged \\
    \midrule
    \multicolumn{3}{l}{\emph{SN15 leaderboard top miner}} \\
    \texttt{agent\_23175}                                 & 77.3\%  & Multi-LLM ensemble; 90-set: ${\sim}65\%$ \\
    \bottomrule
    \end{tabular}
\end{table*}

The pass@$k$ probe on the SFT-stack checkpoint shows pass@$8 = 53.3\%$ versus pass@$1 = 34.8\%$ on the same held-out partition. The roughly 19-point spread says the policy retains substantial latent capability that the supervised stack does not extract at sampling pass@$1$, and is the headroom number Section~\ref{sec:training-grpo} identifies as the target for downstream Dr.~GRPO work.

%

\section{Limitations}
\label{sec:limitations}

The most important limitation of the work reported here is that the model is trained on the smaller of two qualitatively different slices of the SN15 trajectory firehose, and the larger slice is currently outside our training pipeline.

\paragraph{Axis-A and axis-B trajectories.}
Trajectories produced by miner agents on SN15 fall into two qualitatively distinct shapes that we label here for the purposes of the discussion. \textbf{Axis-A} trajectories are agentic: the language model itself emits the \texttt{tool\_call} blocks, the surrounding Python harness only dispatches them, and the policy of where to search next, when to verify, and when to commit to a recommendation lives inside the language model's output. \textbf{Axis-B} trajectories are sub-task: the surrounding Python emits the tool calls (often via deterministic search-strategy code), and the language model is invoked only as a classifier, a match scorer, or a narrator over the dispatch results. The structural-quality filter in Section~\ref{sec:corpus-filter} keeps only axis-A trajectories. This is the right choice for the agentic post-training objective of this paper, because a model fine-tuned on axis-B trajectories acquires the language model's narration and classification roles without acquiring an agent-level policy at all, but it has a consequence we have to name.

\paragraph{The axis-B firehose is the bulk of the data.}
Inspection of recent SN15 race winners (whose source code is open after the post-race embargo) confirms that the top-scoring miner policies on the leaderboard are dominated by sub-task patterns. The race-winning agent whose code we audited in detail at the time of writing is approximately $1{,}850$ lines of Python that calls the language model as a narrow intent classifier, a match scorer, and a recommendation narrator, while routing all of the actual catalogue tool calls in deterministic Python. The leaderboard reliably converges on this pattern because, under the current scoring, it outscores the agentic baseline on every quality axis we have measured. The empirical consequence is that the daily SN15 firehose, in the present configuration of the subnet, is order-of-magnitude dominated by axis-B trajectories: of the roughly $12{,}000$ to $27{,}000$ trajectories produced per day, only the small minority that come from the default agentic baseline or from direct forks of it are axis-A. The trained model in this paper is therefore working from a slice that is much smaller than the published agentic post-training corpora it is being measured against (Glaive Function Calling at approximately $110{,}000$ trajectories, ToolBench at approximately $16{,}000$, ToolACE at approximately $11{,}000$).

\paragraph{What this implies about the headline result.}
The lift we report in Section~\ref{sec:results} is therefore best read as a lower bound on what the subnet trajectory primitive can produce, not as an estimate of the ceiling. The trained model matches the published SFT-only baseline using a corpus that is structurally a small fraction of what the subnet generates on a single day, and the SFT $\to$ RL pipeline that the published recipe runs on top of the corpus has, in our hands, identified the right reward shape (Section~\ref{sec:training-grpo}) but has not yet converted the pass@$8$ headroom into exact-match pass@$1$ within the training budget we ran. The bottleneck is not the recipe family; the bottleneck is the size of the agentic slice the recipe has access to.

\paragraph{Secondary limitations.}
Three further limitations are worth flagging briefly. (i) The held-out numbers reported in Section~\ref{sec:results} are measured through the 7-tool ORO ShoppingBench harness; published baselines are reported in the literature against a 4-tool harness with different system-prompt nudges. We have cross-checked the harness shift on a subset of conditions, but the harness is part of the measurement and a reviewer should not over-interpret a comparison between our numbers and a published number that does not name its harness. (ii) The reasoning-quality coefficient used by the structural filter is itself an LLM judgement, not a gold label. We rely on the coefficient as a per-trajectory process-supervision signal, but it inherits the judge's failure modes; high-judge-quality trajectories are not guaranteed to encode the policy we want to imitate, only to encode reasoning that the judge considers consistent with the recommendation. (iii) The Dr.~GRPO variants we ran do not close the gap to the published SFT+GRPO bar of $48.7\%$. We trace this in Section~\ref{sec:training-grpo} to a combination of reward shape at the turn granularity, importance-sampling drift from the frozen behaviour policy used to seed the rollouts, and the same axis-A versus axis-B skew described above.

%

\section{Future work}
\label{sec:future}

The limitations described in Section~\ref{sec:limitations} map cleanly to two threads of future work.

\paragraph{Improving the RL extraction stage.}
The three Dr.~GRPO variants reported in Section~\ref{sec:training-grpo} together identify a specific failure mode and a clear next step. The third variant established that a dense, per-step, teacher-grounded tool-match reward converts the SFT-stack's latent capability into measurable process improvement within a small number of optimisation steps. The conversion did not reach exact-match success rate on the held-out partition in the twenty steps we were able to run, but the trend was monotonic and consistent with the published recipe's reported convergence over forty to sixty steps. The first downstream step is therefore simply to run the same recipe longer. Beyond that, three structural improvements appear necessary to fully close the pass@$8$ headroom into pass@$1$: a periodic refresh of the rollout policy into the trained policy, which addresses the importance-sampling drift between the frozen behaviour policy that seeds the rollouts and the policy currently being updated; a finer turn-level reward signal that distinguishes good from bad intermediate decisions at the token granularity Dr.~GRPO operates at, rather than rolling up the credit to the rollout level; and a separate process-reward head trained on per-trajectory rule decomposition (price, service, SKU, attribute), which gives the policy a denser gradient on the buckets where the SFT stack already plateaus. Implementations of all three are in progress in our open project and are out of scope for this paper.

\paragraph{Bringing the axis-B firehose into training.}
The much larger lever, however, is on the data side. The current trained model uses only the agentic axis-A slice of the SN15 firehose. The axis-B slice, which is roughly an order of magnitude larger by volume and is currently the dominant shape of leaderboard policy, is unused. We see two paths to bringing it in.

The first path is a corpus-side transformation that synthesises axis-A trajectories from axis-B inputs. The technique is precedented in the published literature for synthetic agentic data (notably ToolACE and the broader family of self-evolution recipes), but the input substrate in our setting is qualitatively different: the axis-B trace already carries a verified outcome, a real catalogue-grounded tool-call timeline, and a reasoning-quality coefficient on the underlying language-model calls. A strong external language model can take the axis-B trace plus the open-source agent code that produced it as joint inputs and rewrite the trace into an agentic axis-A form, with natural-language reasoning between each tool call, grounded in what actually happened. The result is synthetic in the sense that the agentic reasoning is generated, but it is grounded in the sense that the tool calls, the observations, and the final recommendation are all real and verifier-graded. We expect this path to produce, from a single month of subnet trajectories, a corpus of agentic SFT data on the order of the published synthetic corpora it would be compared against, with the additional structural property that the underlying tool calls and observations are not hallucinated.

The second path is on the subnet side, and is the longer-term lever. The reason the axis-B pattern dominates the leaderboard today is that, under the current scoring, an axis-B policy outscores an axis-A policy on the same problem. A modest reweighting of the scoring along an agentic-richness dimension, such that a fraction of the score derives from whether the language model itself emitted the tool calls that the harness dispatched, gradually rebalances the population. We are in the process of designing and pilot-testing this reweighting on the subnet itself, and we expect the resulting shift in the firehose composition to compound with the corpus-side synthesis path described above: the subnet produces more axis-A trajectories at the source, and the corpus-side pipeline rewrites the remaining axis-B trajectories into agentic form. Together, these change the data picture in Section~\ref{sec:limitations} from ``a model trained on the small slice of the firehose'' into ``a model trained on the full firehose,'' which we believe is where the headline number, the pass@$8$ gap, and the recipe-versus-substrate framing of Section~\ref{sec:introduction} all converge.

\paragraph{Reporting cadence.}
Both threads have explicit acceptance criteria internal to our open project, and we intend to publish the next round of results as a focused follow-up: the RL-extraction work paired with a single sharper headline number, and the axis-B utilisation work paired with an analysis of how the recipe behaves as the corpus shifts from axis-A-only to a mixed corpus. The combination, on the timeline we currently project, is what we expect to close the remaining gap to the published SFT+GRPO bar and beyond.

\appendix

\begin{center}
\vspace{0.5\baselineskip}
{\Large\bfseries Appendices}
\vspace{0.2\baselineskip}
\end{center}

%

\section{Corpus construction details}
\label{app:corpus}

\subsection{Trac stages evaluated but not run in production}
\label{app:trac-stages}

Trac's later stages, beyond the converter front-end we use in production, constitute a per-row quality oracle for agentic trajectories. We evaluated all four and chose not to run them at corpus scale on cost grounds. They are summarised here for reproducibility.

\begin{enumerate}
    \item \textbf{Freeze.} A freeze-admit gate that admits only traces whose outcome was successful by default. We added an \texttt{ORO\_ALLOW\_FAILED\_TRACES} flag that admits failed traces too and lets the downstream supervisor decide their usability.
    \item \textbf{Annotate.} Five LLM agents annotate each admitted trace in sequence, each scoring a different dimension (intent understanding, search-strategy quality, observation interpretation, recommendation justification, supervisor pass with a binary \texttt{train\_use\_decision}). Annotators are required to cite specific spans of the trace as evidence.
    \item \textbf{Tasklet compile.} For each accepted trace, Trac compiles a self-contained tasklet whose target product identifiers must be retrievable from policy-visible prompt terms. Tasklets that fail this grounding check are dropped.
    \item \textbf{Autodata verifier.} A teacher-student verifier loop attempts the tasklet with two models of different capability. Tasklets where the teacher succeeds with a calibrated margin over the student, but where the student does not trivially solve the task, are accepted as final training rows.
\end{enumerate}

The end-to-end conversion rate from raw trajectory to a verifier-accepted training row, measured on a representative run, is roughly one in a thousand. The largest single drops are at the freeze stage (most production traces fail strict outcome checks) and the autodata verifier stage (the most common failure mode is that the student model trivially solves the tasklet, which collapses the teacher-student gap). At the corpus scales needed for a ShoppingBench-class model, the annotator cost of running Trac end-to-end on the full firehose is substantial, and only strictly grounded-output frontier models survived Trac's evidence-citation gate.

\subsection{Format-validity gate (five hard rejects)}
\label{app:format-gate}

The format-validity gate hard-rejects any trace that fails any one of five structural invariants.

\begin{enumerate}
    \item \textbf{Tool-call $\leftrightarrow$ tool-response pairing.} Every assistant tool-call must be matched by a corresponding tool-response in the next turn. Unmatched tool calls indicate a truncated or malformed trace.
    \item \textbf{Single \texttt{recommend\_product} call.} The terminal \texttt{recommend\_product} function must appear exactly once across the trace. Multiple recommendations or none disqualify the trace.
    \item \textbf{Parseable recommendation arguments.} The arguments to \texttt{recommend\_product} must parse as valid JSON and resolve to product identifiers that the search server can return.
    \item \textbf{Token-count budget.} The total token count of the trace, measured under the trainer's tokeniser, must be at most $\texttt{max\_length} = 14336$. Longer traces silently truncate during training and are rejected upstream.
    \item \textbf{No assistant-only-think-then-stop.} The trace must not terminate on an assistant think turn without committing to a recommendation. Traces that stop mid-loop without emitting \texttt{recommend\_product} or \texttt{terminate} are rejected.
\end{enumerate}

\subsection{Structural-quality ranking signals}
\label{app:ranking-signals}

Within each problem, candidate trajectories are ranked by a small set of structural-quality signals computed directly from the trace.

\begin{itemize}
    \item \textbf{Think-and-tool-call depth.} A proxy for genuine search refinement rather than one-shot guessing. Higher is better.
    \item \textbf{Search-query reformulation count.} The number of distinct search queries the language model emits. Higher counts indicate iterative refinement.
    \item \textbf{Verification step presence.} Trajectories that issue a \texttt{check\_product\_match} or follow-up \texttt{view\_product} call before the final recommendation are ranked higher.
    \item \textbf{Step-shape regularity.} Trajectories with long stretches of identical tool calls (a proxy for loops, retries, or pathological behaviour) are ranked lower.
\end{itemize}

When two trajectories on the same problem tie on these signals, ties are broken by selecting trajectories with maximally different tool-call shape signatures, so each problem teaches at least two distinct policies wherever possible.

\subsection{Leak-cluster guard for the held-out partition}
\label{app:heldout-construction}

Race-bank problems can be paraphrased, attribute-reordered, or retailer-substituted into surface forms that look distinct but share the same underlying reward specification. The subnet maintains a clustering of problem identifiers by leak-cluster membership, and validators are required to honour the clustering when partitioning race-bank problems between open and held-out. Concretely, we cluster all race-bank problems by leak-cluster identifier, partition leak clusters into a training pool, an evaluation pool, and a never-touch pool, and verify that the intersection between training problem identifiers and evaluation problem identifiers is exactly empty across leak clusters. The evaluation pool is stratified by intent bucket (product, shop, voucher), seeded for reproducibility, and frozen before any training kicks off. The static problem suites are disjoint from both partitions by construction.

\section{Post-training hyperparameters and infrastructure}
\label{app:training}

\subsection{Common SFT configuration}
\label{app:sft-config}

All supervised fine-tuning stages share a common configuration. The base model is Qwen3-4B. We train one epoch with LoRA, sweeping the learning rate over $\{1\!\times\!10^{-5}, 3\!\times\!10^{-5}, 1\!\times\!10^{-4}\}$ and selecting by held-out loss. Across all SFT stages we set $\texttt{max\_length} = 14336$, use SDPA attention, and turn on gradient checkpointing. We use the PrimeIntellect \texttt{renderers} package version 0.1.7 for input rendering and assistant-loss-mask computation; bare Hugging Face \texttt{apply\_chat\_template} for Qwen3-4B produces a token sequence that diverges from the Trac-exported rows by approximately twenty tokens and cannot compute the mask at all because the Qwen template lacks the markers the standard template uses.

\subsection{Practice-run bug catalog}
\label{app:bug-catalog}

The practice run surfaced a handful of bugs that we explicitly guard against in all reported runs.

\begin{itemize}
    \item \textbf{Dropout during generation.} Sampling rollouts under \texttt{model.train()} mode leaves dropout active and corrupts the on-policy data. All rollout sampling is now performed under \texttt{model.eval()}.
    \item \textbf{Loss-mask drift.} The TRL \texttt{DataCollatorForCompletionOnlyLM} silently masks past the assistant boundary on Qwen3 traces because Qwen3 strips the \texttt{</im\_end>} token from non-final assistant turns. We replace it with the PrimeIntellect \texttt{renderers} masker for all SFT stages.
    \item \textbf{Mask validation.} Before every GRPO update we assert that mask$=1$ positions decode to valid assistant content.
    \item \textbf{Stop-token off-by-one.} Qwen3 strips \texttt{</im\_end>} from non-final assistant turns; the rollout-step counter must account for this when computing rewards on the surface text.
\end{itemize}

\subsection{Rejection-sampled re-SFT parameters}
\label{app:rejection-config}

We sample $k = 8$ rollouts per training-set problem at temperature $0.9$, $\texttt{top\_p} = 0.95$, and $\texttt{max\_turns} = 8$ through the multi-turn execution harness. Rollouts are scored under the ShoppingBench bucket-aware reward; rollouts that score $1.0$ are kept. Passing rollouts are concatenated with the SFT corpus at a fixed ratio (50/50 in the reported run, with an ablation against 70/30 noted in the project log) and used as the next-stage training set.

\subsection{Continued teacher-SFT corpus}
\label{app:teacher-config}

The teacher trajectory pool used in this paper comes from running Claude Sonnet 4.6 through the held-out evaluation harness on a $498$-problem race-bank training set. Roughly $210$ of the $498$ trajectories score under production-strict scoring; these are rendered to a step-wise corpus (with the voucher bucket up-sampled $3{\times}$ to match the shop count) and stacked on the rejection-sampled SFT base.

\subsection{KTO configuration}
\label{app:kto-config}

We train KTO with the TRL implementation. The reported runs use $\beta = 0.02$, learning rate $5 \times 10^{-6}$, one epoch, attention-only LoRA with rank $r = 16$, and a context length of $16384$. Training is monitored with an early-stop guard on the \texttt{recommend\_product} emission rate that halts training the moment the rate drops below $0.9$. Final training-time KL is approximately $0.9$. A tighter-$\beta$ variant ($\beta = 0.01$, two epochs, otherwise identical) was trained and evaluated; it over-optimised and regressed across every bucket, and is reported as a negative result.

\subsection{Dr.~GRPO variants}
\label{app:grpo-variants}

We report three Dr.~GRPO variants on top of the post-teacher-SFT checkpoint.

\begin{itemize}
    \item \textbf{v17.} $B = 3$ problems per step, $K = 6$ samples per problem, learning rate $10^{-6}$, KL anchor $\beta = 0.1$, outcome-only reward. Plateaued at $40$-$43\%$ with KL loss spikes into the $17$-$21$ range.
    \item \textbf{v18.} $B = 8$, learning rate $5 \times 10^{-7}$, $\beta = 0.2$, otherwise identical. KL contained within $1.0$-$2.0$; mean reward did not climb.
    \item \textbf{v19.} v18 hyperparameters plus a blended outcome-plus-tool-match reward, $\texttt{tool\_weight} = 0.5$, $\texttt{max\_rollout\_steps}$ raised from $15$ to $25$ to match the held-out harness. Tool-match component grades the policy's per-step tool name and tool argument choice against a Sonnet $4.6$ teacher trajectory. Trained stably for $20$ of a planned $30$ steps on a Modal $12$-hour wall-clock limit. References: \texttt{opd\_sonnet\_500}, $430$ of $527$ training problems matched ($82\%$ coverage).
\end{itemize}

All three variants use the multi-turn execution harness against a deployed Lucene index serving roughly $2.7$M ShoppingBench products. The rollout server runs vLLM $0.9.2$ on a dedicated H200 instance with $\texttt{max\_model\_len} = 32768$ (raised from the previous $16384$ after a context-overflow bug surfaced in v19 with rollout histories averaging $19$ turns).

\bibliography{references}
\end{document}